\documentclass[a4paper, 11pt]{article}
\pdfoutput=1

\newcommand{\ignore}[1]{}
\usepackage{xcolor}
\usepackage{url}
\usepackage{amsmath, amsthm, amssymb}
\usepackage{algorithm}
\usepackage[noend]{algorithmic}
\usepackage{caption}
\usepackage{float}
\usepackage{graphicx}
\usepackage{fullpage}

\usepackage{array}

\newcolumntype{x}[1]{%
>{\centering\hspace{0pt}}p{#1}}%

\begin{document}

\title{Evolving Pacing Strategies for Team Pursuit Track Cycling}
\author{\hspace{-8mm}Markus Wagner, Jareth Day, Diora Jordan, Trent Kroeger, Frank Neumann
\\
    \hspace{-8mm}School of Computer Science\\  \hspace{-8mm}The University of Adelaide, SA, 5005\\
  \hspace{-7mm}$\left[  markus.wagner, jareth.day, diora.jordan, trent.kroeger,  frank.neumann\right]$ @adelaide.edu.au
}
\maketitle

\begin{abstract}
Team pursuit track cycling is a bicycle racing sport held on velodromes and is part of the Summer Olympics. 
It involves the use of strategies to minimize the overall time that a team of cyclists needs to complete a race.
We present an optimisation framework for team pursuit track cycling and show how to evolve strategies using metaheuristics for this interesting real-world problem. Our experimental results show that these heuristics lead to significantly better strategies than state-of-art strategies that are currently used by teams of cyclists.
\end{abstract}

\section{Introduction}\label{sec:intro}
Metaheuristics such as local search, simulated annealing, evolutionary algorithms, and ant colony optimisation have been shown to be very successful problem solvers in various domains \cite{BookTalbi}. This paper describes a novel application of metaheuristics to the optimisation of elite competitive track cycling. Specifically, we examine how they can be used to optimise strategies that lead to improved performance in  the women's team pursuit event. 

Mathematical models of cycling performance have been proposed that aim to predict the outcome of elite single cyclist track and road races~\cite{olds1995modeling,martin2006modeling}. These models balance the estimated anaerobic and aerobic power capabilities of a cyclist against resistive forces such as aerodynamic drag and surface friction in order to determine a velocity profile that corresponds to maximal effort for the cyclist. However, given the non-linear nature of these models, an all-out maximal effort may not necessarily lead to optimal performance. To address this, mathematical optimisation techniques have been applied to the problem of determining an optimal pacing strategy for a single cyclist in track sprint events~\cite{de1999determination}. The work described in this paper builds upon this previous research to propose a framework capable of optimising pacing strategies for a team of cyclists, a problem context in which significantly more complex dynamics are possible.

We present a problem formulation for the team cycling problem and show how to put it into an optimisation framework. The framework takes into account that cyclists upfront spend more energy than cyclists that are in the slipstream of others. This implies that cyclists have to change their positions within the team during the race in order to minimize the overall time. In our optimisation problem, we have to determine when such changes occur as well as the power applied by the whole team during different parts of the race.

Our problem has some relation to scheduling problems. Scheduling problems have been addressed in operations research and computer science for several decades. Recent examples of real-world scenarios where evolutionary techniques were applied are, amongst others, vehicle scheduling in public transport~\cite{bunte2009}.
Our problem has similarities to the scheduling of jobs with controllable processing times~\cite{DBLP:journals/dam/ShabtayS07} as we have to determine at each point of time the power that the team applies in the race. However, it involves more difficulties as we also have to determine a strategy for changing the order of the riders.

From an algorithmic perspective, the problem is interesting and challenging as it requires the simultaneous optimisation of discrete and continuous variables. The discrete variables determine the scheduling of the different cyclists in the lead position during a race, whereas the continuous variables determine the level of power applied by that cyclist. We present different approaches for dealing with this mixed discrete and continuous setting and evaluate the quality of the different approaches with respect to standard strategy and power settings. Our results show that state-of-the art strategies for team pursuit track cycling can be significantly improved using metaheuristic approaches.

This paper is structured as follows. Section~\ref{sec:problem} outlines the specific problem to be solved in detail, with its formulation as an optimisation problem following in Section~\ref{sec:formulation}. In the subsequent Section~\ref{sec:ea}, our evolutionary approach to this problem is described, with results being presented and discussed in Section~\ref{sec:experiments}. The paper concludes in Section~\ref{sec:conclusions} with a summary of key findings and a description of potential areas for future research.


\section{Team Pursuit Track Cycling}\label{sec:problem}

Track cycling is raced on elliptical tracks called velodromes. These can be indoor or outdoor and are usually 250 metres in circumference, which includes two straight sections and two banked turns. Track cycling comprises a number of sprint and track endurance events. Both individual and team events exist. These include pure sprint events such as the individual and team sprints; long sprints such as the 500 m and 1000 m time trials and the \emph{Kieren}; middle distance events such as the individual and team pursuits; and endurance events such as the \emph{Madison}, scratch, handicap and points races. As described in the previous section, mathematical models of performance have been developed for predicting and optimising a subset of the individual events. 
This paper focuses on the application of a specific metaheuristic to optimise both the pacing strategy, in terms of the power applied by each cyclist in first position, and the transition strategy employed within the context of a women's team pursuit event.

The team pursuit is an event in which multiple cyclists work together to complete a given race distance in the minimum possible time. The term `pursuit' is derived from the fact that races begin with two opposing teams that begin on opposite sides of the track and try their best to pursue each other. Team members take turns riding in the front position thus allowing the other team members to draft closely behind for maximum aerodynamic benefit. This rotation allows the team to maintain a significantly higher velocity than would be achievable by a single cyclist. Changes to the relative position of the cyclists are most efficiently performed on either of the two banked turns of the track, where the cyclist in first position sweeps up the bank of the turn and then rejoins in last position. For the men's team pursuit event, there are four cyclists in each team and the race consists of sixteen laps for a total of 4000 metres. In this event, only three of the four cyclists need to complete the race and so it is common to see one cyclist sacrifice themselves for the team by spending longer than average in first position to the point where his energy reserves are exhausted and he leaves the race. In the women's team pursuit event, there are three cyclists in each team and the race is over twelve laps for a total of 3000 metres. Unlike the men's variant, all three cyclists in the women's team pursuit must complete the entire race distance, which typically leads to a more even distribution of workload amongst the team. The standard transition strategy observed from world championship and Olympic women's team pursuit events usually involves cyclist transitions occurring after 187.5 metres (0.75 of a lap) and then every 250 metres (1 lap) until the 2562.5 metre point (1.25 laps from the finish), after which no further transitions occur.

Due to its lower complexity, the women's team pursuit event was chosen as the basis for the work described in this paper. There are four main ways by which to improve performance in the women's team pursuit event. The first is to improve the physiological and psychological capabilities of the cyclists through training. The second is to improve the technical specifications of the bicycles used, in terms of their mass and aerodynamic properties. The third is to change the pacing strategy used by the team such that the power applied results in maximal benefit. Finally, the transition strategy may be changed to achieve a more even distribution of energy usage by the team and the most aerodynamically effective velocity profile. This paper will focus on the application of metaheuristic algorithms to optimise both the pacing strategy, in terms of the power applied by each cyclist in first position, and the transition strategy employed within the context of a women's team pursuit event. This problem is interesting in that it requires the simultaneous optimisation of both discrete and continuous variables. The discrete variables in this problem identify the specific positions within the race where a transition occurs. These transition points are constrained to occur only on the two banked turns of the track and may therefore be represented by integer variables that identify half-laps throughout the race. The continuous variables describe the power to be applied by the cyclist in first position between each transition, which may vary continuously between zero and the maximum power within the physiological capability of the cyclist. The non-linear nature of the physiological and kinematic cycling model used for this project means that it is difficult, if not impossible, to frame this problem in terms of traditional mathematical optimisation approaches and so we have developed a hybrid metaheuristic optimisation framework that simultaneously optimises these discrete and continuous variables.


\section{Problem Formulation}\label{sec:formulation}
The application of evolutionary algorithms to numerical optimisation problems has become a common approach. 
In our case of designing good race strategies, the high-level optimisation of the cyclists' tasks has to be combined with an efficient floating-point optimisation for the individual sections of a strategy.

For the problem at hand, we propose an evolutionary algorithm that is based on ideas taken from the field of genetic algorithms~\cite{holland-75} and evolution strategies~\cite{Rechenberg1973}. 
We implemented our algorithms in the jMetal framework \cite{DurilloNA10}, and the code is available upon request.

In Section~\ref{sec:problem}, we discussed the nature of team pursuit cycling races, and noted that the aim of our optimisation was the minimisation of the race time. We determined that the core parameters to be optimised are the number of half-laps the first rider performs before the cyclists change position and the power output of the first rider. We thus have two sets of parameters that we need to form a solution for the race model, the half-laps before rider transition and the first rider power output per lap. 

A transition moves the first cyclist to the rear and the second to the front. These transitions occur exclusively on one of the two banked turns and so there is a minimum of one half-lap between transitions. The power output of the first rider per half-lap is able to be simplified in that a rider should maintain a certain power level at all times while in the lead. Due to this, only one power output value is needed per half-lap before transition. We aim to maintain a fixed number of variables for simplicity; with the worst case of riders transitioning every half-lap we find that it is necessary to store a number of variables equal to twice the number of half-laps being represented.

While the half-laps before transition can simply be represented by integers, the power output is measured in $W$ and must be a floating point number to correctly capture precision. Due to both integer and floating-point values represented in the solution, we choose a floating-point-valued representation in order to maintain a simple, fixed-length solution without sacrificing precision in the manner discussed by Herrera et al.~\cite{herrera1998tackling}

\begin{figure}[h]
	\centering
	\includegraphics{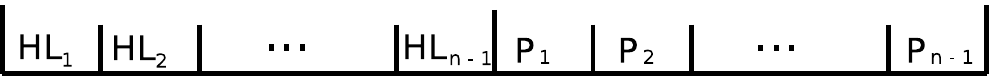}
	\caption{Illustration of the solution. $\mathit{HL}$ denotes the half-laps before rider transition and $P$ indicates power output per first rider per half-lap.
		\label{fig:representation}
	}
\end{figure}

In section \ref{sec:problem} we noted that there is typically an initial 0.75 laps, or 1.5 half-laps, required at the start of the race before transitions can be made, and a final 1.5 half-laps at the end of the race in which transitions may not be made. In this way, these initial and final race segments are atomic and are hence represented within the solution in the same manner as regular half-laps. Rather than representing $n$ half-laps, we instead represent $n-2(1.5)+2$. As we are storing two sets of information based on the number of transitions set to the maximum of $n-1$, the total number of parameters in the representation is $2(n-1)$. Figure \ref{fig:representation} illustrates the solution.

If any half-laps before transition is greater than one, there will be extraneous information in both halves of the solution. We are thus interested only in the first $m$ variables on both halves of the solution, where $m$ is derived from $(\sum_{k=1}^{m}\mathit{HL}_k)\ge (n-1)$ and $\mathit{HL}_k$ is the $k^{th}$ element of the half-laps before transition set. To illustrate, given a solution with the half-laps before transition set equal to [3, 3, 3, 3, 3, 3, 3, 3, 3, 3, 3, ...] and $n=24$, we would only be interested in the first eight half-laps before transition and the corresponding eight power levels since $(\sum_{k=1}^{8}\mathit{HL}_k) = 24 \ge 23$.

\subsection{Team Pursuit as an Optimisation Problem}

The fitness evaluation function uses time as a measure of fitness while making sure that the energy expended for each rider throughout the race is within the physiological capability of that rider. In order to calculate the given time for a race we perform two tasks. First, we modify the model taken from \cite{martin2006modeling} in order to be suitable for the given problem. This then provides the formulae to calculate the power for the riders to ride a given distance. Second, we use forward integration modelling in order to predict the time taken using the solution being evaluated. Both tasks are set into context in pseudo-code form in Algorithm~\ref{alg:fitnessFunction}, and are explained in greater detail in the following.

\begin{algorithm}[H]
\renewcommand{\algorithmicrequire}{\textbf{Input:}}
\renewcommand{\algorithmicensure}{\textbf{Output:}}
\caption{Fitness Function: Evaluating a Pacing Strategy} 
\label{alg:fitnessFunction} 
\begin{algorithmic}[1]
\REQUIRE list of transitions $transitions$, list of associated power levels $powers$
\ENSURE the predicted race time $totalTime$
\STATE set $\mathit{rider}_1$, $\mathit{rider}_2$, and $\mathit{rider}_3$ according to their preset initial order
\STATE $\mathit{totalTime} := 0$
\FOR[iterate in order over the transition strategies]{$i \in \left\{ 1, \dots, s.length\right\}$}
	\STATE $\mathit{h} := \mathit{transitions}\left[i\right]$ 
		\COMMENT{get the current number of half laps}
  \STATE $\mathit{d} := \mathit{h} \cdot \mathit{halfLapLength}$
	  \COMMENT{set the current distance rode}
	\STATE $\mathit{p} := \mathit{powers}\left[i\right]$ 
		\COMMENT{get the current power level}
	\STATE $\mathit{t} := \mathit{forwardIntegration}(\ldots)$
	  \COMMENT{compute the time needed for cycling $d$, using Equation~\eqref{eq2}}
	\STATE $\mathit{totalTime}\ +\hspace{-1mm}= t$
	  \COMMENT{increase the race time by the time $\mathit{rider}_1$ needs to cycle the distance $d$}
	\STATE $\mathit{rider_1.totalEnergy}\ -\hspace{-1mm}= \mathit{p} \cdot \mathit{t}$
	  \COMMENT{reduce the first rider's energy}
	\STATE $\mathit{useEnergy(rider_2)}$ \& $\mathit{useEnergy(rider_3)}$
	  \COMMENT{reduce the other riders' energies using Equation~\eqref{eq1}, based on them keeping up with $\mathit{rider}_1$'s speed} 
	\STATE $\mathit{swapRiderConfig()}$ 
	  \COMMENT{reorder the riders}
	\STATE $\mathit{totalTime}\ +\hspace{-1mm}= transitionTime$
	  \COMMENT{increase the race time by the time lost by reordering the riders}
\ENDFOR
\end{algorithmic}
\end{algorithm}

\subsubsection*{Task 1: Modified Model for Team Pursuit}
The model taken from \cite{martin2006modeling} was adapted to suit track cycling conditions. This involved removing the potential energy as vertical elevation is negligible, and combining air speed and ground speed as the cycling is performed in still air conditions. This gives us the following formula:

\begin{equation}\label{eq1}
    \Delta \mathit{KE} = (P \times E - C_D A \times (\frac{1}{2} \rho \textrm{v}^3) - \mu \times (\textrm{v}F_N)) \times \Delta t 
\end{equation}

in which $\mathit{KE}$ is kinetic energy, $P$ is power, $E$ is mecahnical efficiency, $C_D A$ is the frontal area of the bike, wheel spokes and rider, $\rho$ is air density, v is speed, $\mu$ is a global coefficient of friction, $F_N$ is the weight of the bike and the rider, and $t$ is time.
We then calculate the acceleration over a given distance, which is obtained from the number of laps in a given configuration for the current solution, by assuming that the excess kinetic energy is used for this acceleration. The acceleration is then used to calculate the time taken for the rider to cover the given distance.

As the second and third rider have a drafting benefit from being in the slipstream of the first rider, we use the following formula\footnote{It is an extension based on \cite{martin2006modeling} to incorporate drafting benefits.} to calculate the power used to keep up with the acceleration of the first rider:

\begin{equation}\label{eq2}
    P = (C_D A \times C_{Draft} \times \frac{1}{2} \rho \textrm{v}^3 + \mu  \times (\textrm{v}F_N) + \frac {\Delta \mathit{KE} }{ \Delta t} )/E
\end{equation}

in which $E$ is the efficiency of the drive system, and where $\Delta \mathit{KE}$ is calculated using the final velocity of the first rider after acceleration. $C_{Draft}$ is drafting coefficient of the rider and represents the reduction in the $C_D A$ of a cyclist due to the aerodynamic benefits of drafting in second or third position. 

\subsubsection*{Task 2: Forward Integration Modelling for Race Time Computation}
The algorithm steps through the different configurations the cyclists would transition to throughout the race. When the riders transition, it is assumed to take a fixed amount of time which is added to the total time of the race. For each of the different configurations, the number of half laps rode in the current configuration is used to calculate the distance rode. The time taken to ride this distance is then calculated by forward integration, using Equation~\eqref{eq1}, and a small $\Delta t$ (0.1 seconds). This time is then added to the total race time. The energy used to ride the distance in this configuration is subtracted from the total energy of the first rider, in order to ensure the rider does not run out of energy.  This total energy is obtained from the initial amount of available energy each rider is assumed to have, which can be arbitrarily allocated throughout the race within physical limitations. Equation~\eqref{eq2} is then used to calculate the energy used by the second and third rider in order to maintain pace with the first rider. When the cumulative sums of the distances cover the race distance, the time taken is then used as a measure of the fitness of the solution currently being evaluated. 

\section{Metaheuristic Approach}\label{sec:ea}

With the use of a powerful optimiser for the first part, we could concentrate our efforts on finding transition strategies. The evaluation of a single race strategy is the race time of its transition strategy, for which an optimised power profile is computed with respect to the given fitness function.

We use a state-of-the-art evolutionary algorithm as our local optimiser for the power profiles called Covariance Matrix Adaptation Evolution Strategy (CMA-ES)~\cite{hansen2006eda}. 
It self-adapts the covariance matrix of a multivariate normal distribution. This normal distribution is then used to sample from the multidimensional search space where each variate is a search variable. The covariance matrix allows the algorithm to respect the correlations between the variables making it a powerful evolutionary search algorithm. 

Its internal management of perturbation range and other parameters allows it to overcome problems such as parameter tuning, poor performance on badly scaled fitness functions, the need to scale population sizes in proportion to problem dimension, and premature convergence of the population.

In our setup, the initial population of CMA-ES's optimisation process is set close to the power levels of the currently best solution $x^*$, with the initial standard deviation being $\pm$10 W. This way, time is saved to arrive in the right order of magnitude regarding the power levels, compared to starting with randomly initialized solution vectors. 

During the power level optimisation, constraint violations are handled by the fitness function, which adds high weights to the predicted race time in case a rider runs out of energy or the power levels exceed the limits.

Later on, in Section~\ref{sec:ppo}, we will briefly compare the performance of CMA-ES as our local optimiser to a simple random search variant, where the latter turns out to be inferior and even counterproductive regarding the resulting race time.\footnote{Note that other optimiziation software may outperform CMA-ES, but as its algorithmic setup is virtually parameter-free, no fine-tuning of parameters was required.}

\subsection{Transition Strategy Optimisation}

When we constructed our algorithms for the strategy optimisation, we had to consider that the above-described continuous power profile optimisation is relatively time-consuming, with each evaluation being a whole run of CMA-ES, and each taking several seconds. 
We therefore chose to use random local search (RLS) and a simple evolutionary algorithm (Simple EA) variant, as described below. Both use only a population size of $\mu=1$, and we conjecture that larger population sizes are not necessarily beneficial in the face of the significantly increased runtime of the algorithm.

Note that our applied operators first determine the effective length $m$ of a strategy, i.e. the subsequence (starting at the first entry) that is sufficient to finish the race (as described in Section~\ref{sec:formulation}).

For RLS, the neighbourhood is defined as follows: given the current transition strategy $s$, the neighbourhood contains all those strategies $s'$ that can be reached by incrementing (respectively decrementing) a single field in the strategy. Note that this can potentially change the strategy's effective length. However, this is unproblematic because the individual's length is sufficient to guarantee a valid strategy to complete the race.

The mutation operator used in the Simple EA randomly changes the value of a field of the effective strategy to a different value. 
Whenever a new individual is created, the expected number of such changes is one, as the probability of performing such a change is $p=1/m$. Thus, contrary to RLS, several positions can be modified during one generation of an offspring.

\renewcommand{\algorithmicloop}{\textbf{repeat forever}}
\begin{figure}[h!]
\caption{The algorithms used for the transition strategy optimisation. The fitness function $f$ evaluates a given strategy $x$ by finding an optimised power profile for it, and then returning the race time based on $x$ and the profile.\vspace{-0.3cm}}
 \begin{minipage}[t]{7.5cm}
 \begin{algorithm}[H]
\caption{Random Local Search} 
\label{alg:rls} 
\begin{algorithmic}[1]
\STATE Choose $x^*$ uniformly at random.
        \LOOP
        		\REPEAT
        		\IF{there is an unseen neighbour of $x^*$} 
        			  \STATE Create said neighbour $x$ of $x^*$.
        		\ELSE
								\STATE End of algorithm.
						\ENDIF
        		\UNTIL{$f(x) \le f(x^*)$}
        		\STATE $x^*:=x$
       \ENDLOOP
\end{algorithmic}
\end{algorithm}
 \end{minipage}
 \hfill
 \begin{minipage}[t]{7.8cm}
\begin{algorithm}[H]
\caption{Simple EA} 
\label{alg:simpleea} 
\begin{algorithmic}[1]
\STATE Choose $x^*$ uniformly at random.
        \REPEAT
     			  \STATE Create $x$ by mutating each position of $x^*$ independently with probability $1/m$.
        \IF{$f(x) \le f(x^*)$} 
        		\STATE $x^*:=x$
        \ENDIF
       \UNTIL{maximum number of fitness evaluations reached.}
\end{algorithmic}
\end{algorithm}
 \end{minipage}
\end{figure}


\section{Experiments}\label{sec:experiments}
\subsection{Experimental Parameters}
To make a realistic model of the race, we have implemented a number of parameters as listed in Table~\ref{tab: Race characteristics}. Of note, time to transition is the time lost when the riders transition. We approximate the power a rider can output by their personal mass multiplied by 5 $Wm^{-1}$, and determine that the available energy is this power over 210 seconds. We use a standard 8 kg for a bicycle's mass. $C_D A$ is the effective frontal area of the rider. Finally, we impose an upper bound of 1000 $W$ and a lower bound of 100 $W$ to ensure a feasible solution, and model fatigue by implementing a maximum number of half-laps before a rider has to transition.

\begin{table}[ht]
\caption{Race characteristics\vspace{-0.3cm}}
\begin{center}
\small{
\begin{tabular}{|c|c||c|c|}
\hline
Gravitational acceleration & 9.80665 $ms^{-2}$ & Rider A mass & 70.0 $kg$ \\\hline
Mechanical efficiency & 0.977 & Rider B mass & 67.5 $kg$ \\\hline
Global friction & 0.0025 & Rider C mass & 65.0 $kg$ \\\hline
Temperature & 20$^{\circ}$ C & Available energy & mass $\times$ 5 $\times$ 210 $J$ \\\hline
Air pressure & 1013.25 $hPa$ & Bicycle mass & 8.0 $kg$ \\\hline
Relative humidity efficiency & 50\% & Rider A $C_D A$ & 0.190 \\\hline
Race distance & 3000 $m$ & Rider B $C_D A$ & 0.175 \\\hline
Maximum power & 1000 $W$ & Rider C $C_D A$ & 0.160 \\\hline
Minimum power & 100 $W$ & $C_{Draft}$ second position & 0.7 \\\hline
Maximum half-laps before transition & 3 & $C_{Draft}$ third position & 0.6 \\\hline
Laps & 12 & Time to transition & 0.12 $s$ \\\hline
\end{tabular}}
\end{center}
\label{tab: Race characteristics}
\end{table}

\subsection{Power Profile Optimisation}\label{sec:ppo}
In order to demonstrate the fitness evaluation, and to form a basis for comparison with optimised solutions, we investigate a set transition strategy: the first rider performs the initial 0.75 laps, and then the riders transition every two half-laps. This transition strategy is represented as [1, 2, 2, 2, 2, 2, 2, 2, 2, 2, 2, 2]. We use the initial rider configuration [A, B, C], allowing the riders with the least energy to gain benefit from drafting. To increase readability, we use the simpler notation of three capital letters instead of the list style.

We investigate two different power profiles for this demonstration. The first profile, Unoptimised 1, has the first rider output a greater amount of power in order to reach a higher velocity by the end of the initial 0.75 laps, but upon transition has the riders drop back to a lower power for the rest of the race. We use the arbitrarily high value of 900 $W$ with a lower constant of 364 $W$, the highest that can be used while maintaining positive rider energy.

The second profile, Unoptimised 2, uses a constant power throughout the race, sacrificing initial velocity for a smoother acceleration. We use 409 $W$, the highest constant power value possible without entering negative energy values by the end of the race. Table \ref{tab:unoptimised} illustrates the race time and remaining energy for the two solutions with the initial rider configuration ABC.
 
In addition, we created an algorithm that generates a solution iteratively by cycling through the same transition strategy, attempting to add or subtract power from the power profile in order to minimize the remaining energy of all the riders at the end of the race. We again use the initial rider configuration ABC, and create two different power profiles. The first solution, Iterative 1, aims for a high energy output for the first 0.75 laps, and the second, Iterative 2, aims for an even power distribution. Table \ref{tab:unoptimised} shows the race time for the resulting solutions. The first iterative solution has the power profile [924, 350, 393, 324, 350, 393, 324, 350, 393, 324, 350, 393], and the second iterative solution has the power profile [373, 355, 460, 373, 355, 460, 373, 355, 460, 373, 355, 460].

\begin{table}[ht!]
\caption{Resulting times on unoptimised and iterative solutions with the initial rider configuration ABC.\vspace{-0.3cm}}
\begin{center}
\begin{tabular}{|x{0.17\textwidth}|x{0.17\textwidth}|x{0.17\textwidth}|x{0.17\textwidth}|x{0.17\textwidth}|}
\hline
Power profile & Race time (s) & Rider A final energy (J) & Rider B final energy (J) & Rider C final energy (J) \tabularnewline\hline\hline
Unoptimised 1 & 208.42 & 39.89 & 1294.01 & 1861.89 \tabularnewline\hline
Unoptimised 2 & 209.92 & 21.50 & 549.97 & 1657.38 \tabularnewline\hline\hline
Iterative 1 & 208.92 & 57.06 & 30.10 & 65.18 \tabularnewline\hline
Iterative 2 & 213.22 & 28.57 & 26.67 & 48.36 \tabularnewline\hline
\end{tabular}
\end{center}
\label{tab:unoptimised}
\end{table}

We can see from Table \ref{tab:unoptimised} that iteratively altering the power to maximize rider power output will not necessarily return a better solution. Although the riders have less energy at the end of the race in the iterative solutions, the best iterative solution is slower than the best unoptimised solution.

We next run this set transition strategy under CMA-ES in order to determine if we can gain a faster race time by optimising the power profile. We use CMA-ES on this single solution for 100 repetitions, each with 2000 evaluations. Table~\ref{tab:setStratoptimisedABC} lists the results for the optimised strategy.

\begin{table}[ht]
\caption{Resulting time on optimised solution with the initial rider configuration ABC.\vspace{-0.3cm}}
\begin{center}
\begin{tabular}{|x{0.14\textwidth}|x{0.14\textwidth}|x{0.14\textwidth}|x{0.14\textwidth}|x{0.14\textwidth}|x{0.14\textwidth}|}
\hline
Best race time (s) & Mean time (s) & Standard Deviation & Rider A final energy (J) & Rider B final energy (J) & Rider C final energy (J) \tabularnewline\hline\hline
204.52 & 205.403 & 0.482 & 35 & 96 & 1948 \tabularnewline\hline
\end{tabular}
\end{center}
\label{tab:setStratoptimisedABC}
\end{table}

\begin{figure}[ht]
\begin{minipage}[b]{0.45\linewidth}
\centering
\includegraphics[scale=0.55]{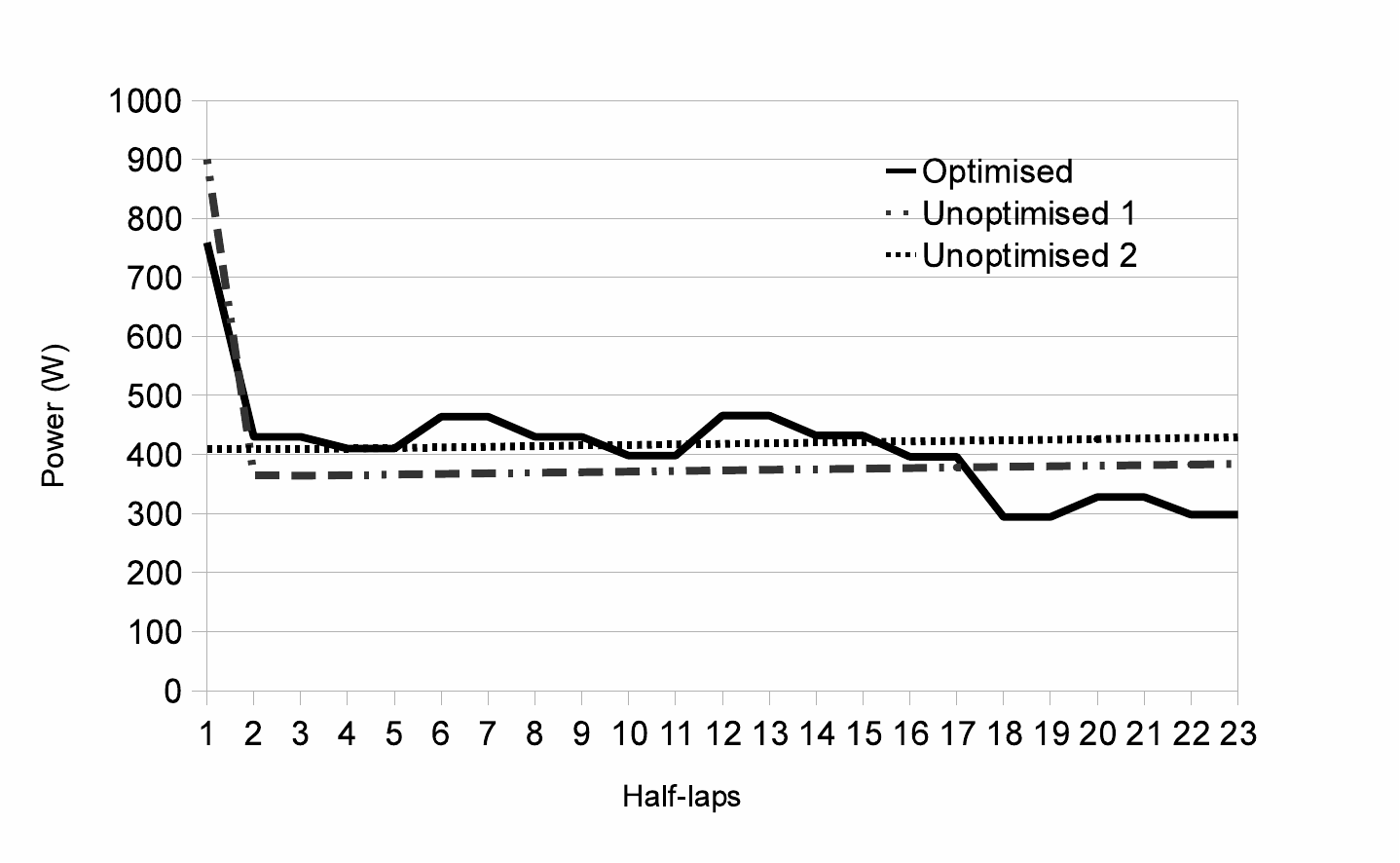}\vspace{-0.3cm}
	\caption{Comparison of the optimised solution's power profile to that of the two unoptimised solutions, all under the same strategy.
		\label{fig:powerProfileStatic}
	}
\end{minipage}
\hspace{0.45cm}
\begin{minipage}[b]{0.5\linewidth}
\centering
\includegraphics[scale=0.55]{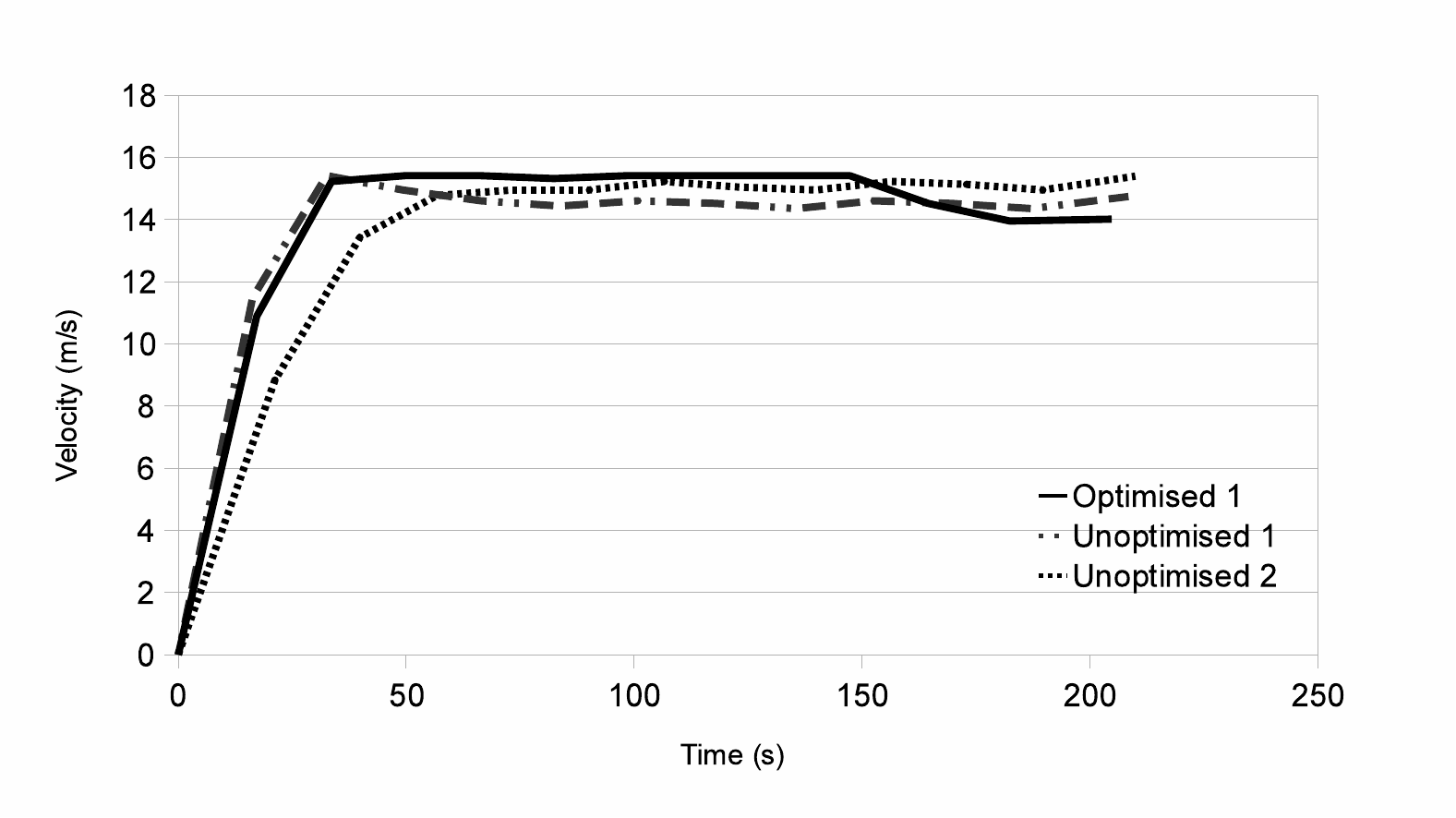}\vspace{-0.3cm}
	\caption{Comparison of the optimised solution's velocity profile to that of the two unoptimised solutions, all under the same strategy.
		\label{fig: VelocityProfileStatic}
		}
\end{minipage}
\end{figure}

This optimised solution uses the same transition strategy with the power profile [759, 430, 410, 464, 430, 398, 466, 432, 396, 294, 328, 298]. It returns a best time of 204.52, 3.90 seconds better than the best unoptimised time. Figures~\ref{fig:powerProfileStatic} and \ref{fig: VelocityProfileStatic} compare the power and velocity profiles of the optimised and the two unoptimised solutions.

In addition to the initial ABC rider configuration, we are able to extend the model by investigating the optimisation of all permutations of the riders. Table~\ref{tab:setStratoptimisedAll} investigates the optimised solution with the selected strategy, based on the other permutations of the initial rider configuration. This is again performed with 100 repetitions and 2000 evaluations per repetition.

\begin{table}[ht]
\caption{Resulting time on optimised solution with all other configurations\vspace{-0.3cm}}
\begin{center}
\begin{tabular}{|c|c|c|c|}
\hline
Configuration & Best race time (s) & Mean time (s) & Standard Deviation \\\hline\hline
ABC & 204.52 & 205.403 & 0.482 \\\hline
ACB & 204.82 & 205.608 & 0.486 \\\hline
BAC & 204.82 & 205.507 & 0.394 \\\hline
BCA & 203.32 & 204.006 & 0.383 \\\hline
CAB & 205.22 & 205.947 & 0.452 \\\hline
CBA & 204.32 & 204.830 & 0.327 \\\hline
\end{tabular}
\end{center}
\label{tab:setStratoptimisedAll}
\end{table}

We can see that we are able to return a better time by altering the initial configuration of the riders. Here, BCA gives a better time of 203.32 with the power profile [760, 404, 467, 432, 395, 467, 435, 393, 471, 411, 385, 126]. This is interesting, as it may be intuitive to predict that placing the rider with the most energy in the initial high power phase would be optimal. It is also interesting to note the low standard deviation, which shows a relatively consistent race time is returned.

\subsection{Strategy Optimisation}
After demonstrating the decreased race time gathered from the power optimisation, we now demonstrate combined transition strategy and power optimisation. Running RLS and Simple EA over the six permutations of initial rider configuration returned the results in Table \ref{tab:completeOptimised}. These results were obtained by performing 100 repetitions each with 2000 CMA-ES evaluations. The Simple EA was run for 100 EA evaluations.

\begin{table}[ht]
\caption{Optimising transition strategy and power profile for all rider configurations\vspace{-0.3cm}}
\begin{center}
\begin{tabular}{|c|c|c|c|c|}
\hline
Algorithm & Configuration & Best race time (s) & Mean time (s) & Standard Deviation \\\hline\hline
 
RLS & ABC & 202.08 & 202.881 & 0.864 \\\hline
RLS & ACB & 202.08 & 202.861 & 0.628 \\\hline
RLS & BAC & 202.20 & 202.841 & 0.507 \\\hline
RLS & BCA & 201.96 & 202.945 & 0.712 \\\hline
RLS & CAB & 202.20 & 203.180 & 0.829 \\\hline
RLS & CBA & 202.10 & 203.198 & 0.837 \\ \hline \hline

Simple EA & ABC & 201.98 & 203.084 & 0.656 \\\hline
Simple EA & ACB & 202.20 & 203.101 & 0.610 \\\hline
Simple EA & BAC & 201.90 & 203.203 & 0.683 \\\hline
Simple EA & BCA & 202.02 & 202.688 & 0.385 \\\hline
Simple EA & CAB & 202.20 & 203.049 & 0.542 \\\hline
Simple EA & CBA & 202.22 & 203.087 & 0.614 \\\hline

\end{tabular}
\end{center}
\label{tab:completeOptimised}
\end{table}

\begin{figure}[ht!]
\begin{minipage}[b]{0.45\linewidth}
\centering
\includegraphics[scale=0.6]{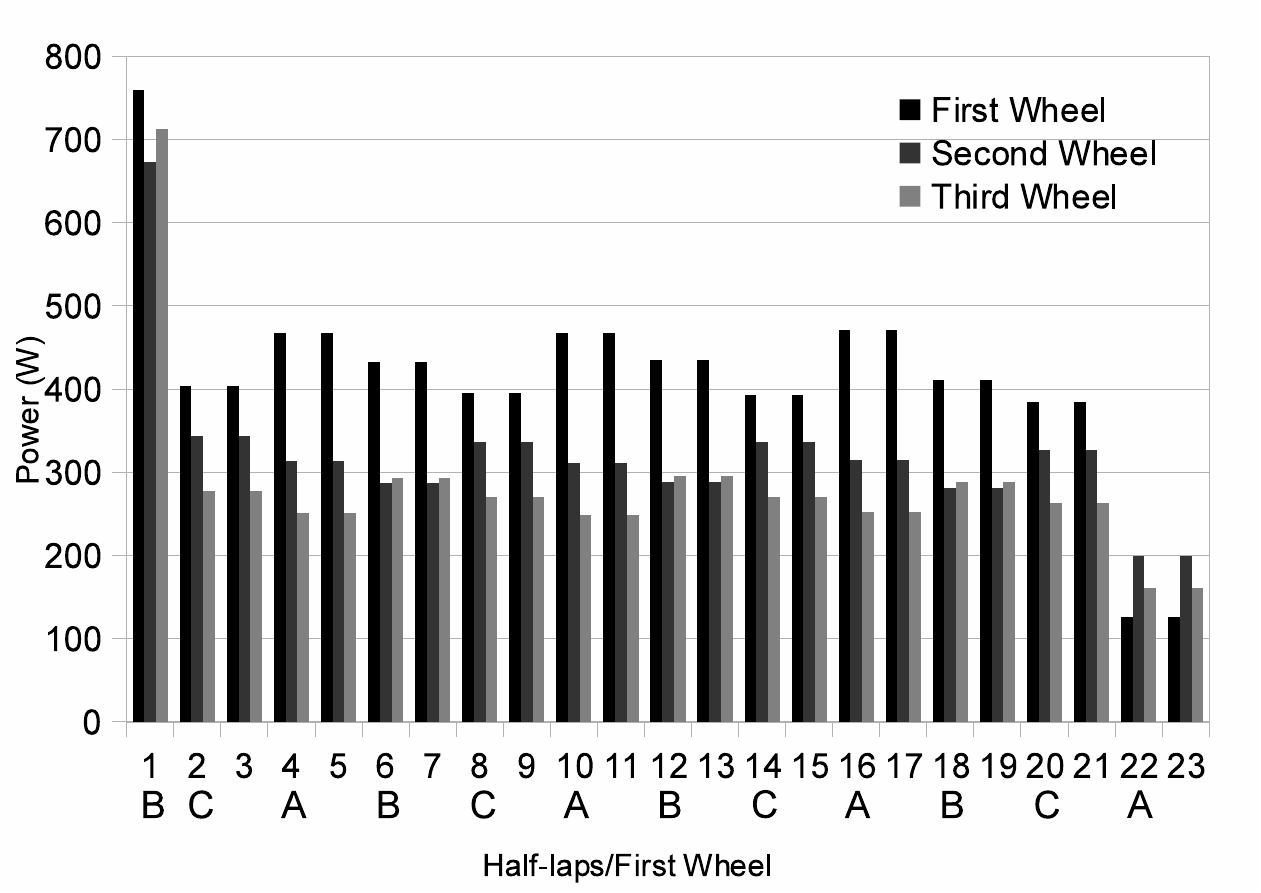}
	\caption{The power profile of the best power-optimised solution, BCA.
		\label{fig:PowerProfilePowerOptimised}
	}
\end{minipage}
\hspace{0.45cm}
\begin{minipage}[b]{0.5\linewidth}
\centering
\includegraphics[scale=0.6]{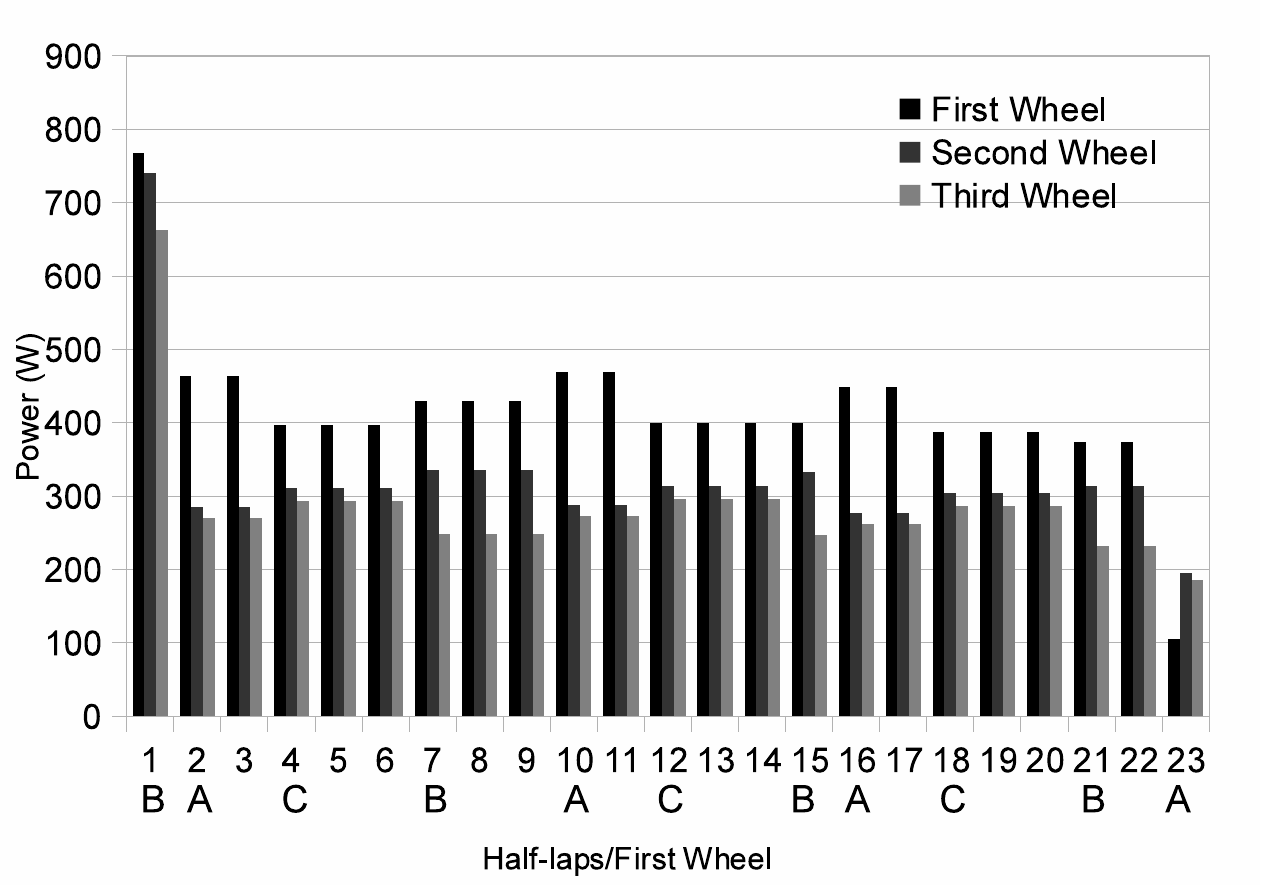}
	\caption{The power profile of the best strategy-optimised solution, BAC under Simple EA.
		\label{fig:PowerProfileStrategyOptimised}
		}
\end{minipage}

\end{figure}

\begin{figure}[ht!]
\centering
\includegraphics[scale=0.55]{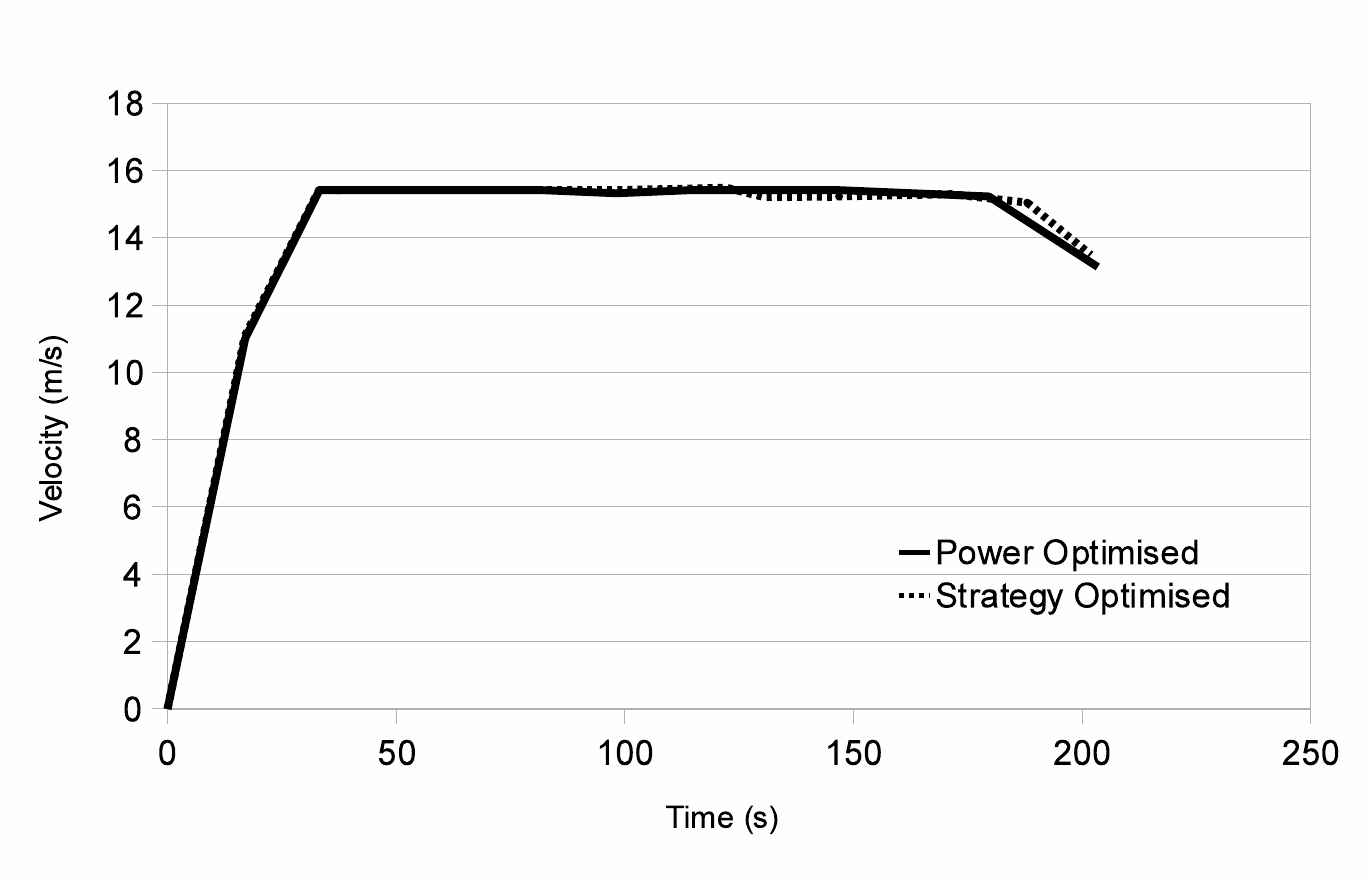}\vspace{-0.3cm}
	\caption{Comparison of the velocity profiles of the power-optimised BCA (203.32$s$) and the Simple EA strategy-optimised BAC (201.90$s$).
		\label{fig:OptimisedVelocity}
		}
\end{figure}

We can see that an evolutionary approach to the transition strategy and power profile is able to further speed the race, allowing consistent times of 202 seconds. The BAC-based strategy evolved by the Simple EA with the transition strategy [1, 2, 3, 3, 2, 3, 1, 2, 3, 2, 1] and the power profile [767, 464, 397, 430, 470, 400, 400, 449, 387, 373, 105] is able to reach the best time of 201.90 seconds, 1.42 seconds faster than the solely power-optimised BCA. Figures \ref{fig:PowerProfilePowerOptimised} and \ref{fig:PowerProfileStrategyOptimised} demonstrate the power profile of all three riders for both the power-optimised BCA and the strategy-optimised BAC. Figure \ref{fig:OptimisedVelocity} compares the two velocity profiles; it is interesting to note how little the two profiles vary for a 1.42 seconds difference in time.


\section{Conclusions and Future Work}\label{sec:conclusions}

In this paper, we addressed the problem of evolving team strategies for track cycling. We introduced a model of a team consisting of cyclists with different properties, such as weight, maximum power output, and available amount of energy. This allowed us to evaluate strategies without having athletes explicitly evaluating the strategies on track.

The experiments revealed several interesting insights. It is important to note that it is undesirable to manually generate the power profile for a race, and that evolutionary algorithms may be used to reduce a race time by optimising the power profile. However, all permutations of initial rider configuration should be investigated because it may not be obvious which configuration of riders is optimal for a particular race. Finally, an algorithm that optimises both transition strategy and power profile is able to obtain a superior race time. The RLS and Simple EA algorithms were able to return a considerably faster time than either the unoptimised or transition-strategy optimised solutions for the domain.

In the future, we will continue our research in the following areas:
\begin{itemize}
	\item Validation and improvement of the obtained solutions through cooperation with professional cyclists.
    \item Investigation into the sensitivity of the obtained solutions to parameter variations.  As there may be variations or errors in parameter measurement, it would be of interest to produce a stochastic model that gives the confidence range of a given solution.  An additional objective would be to then reduce the standard deviation of the stochastic model's distribution of results, thus creating a multi-objective optimisation, as we want a low race time but also high confidence.
\end{itemize}

\subsection*{Acknowledgement}

The authors would like to thank Dr David Martin from the Australian Institute of Sport cycling program and Dr Tammie Ebert from Cycling Australia for their valuable support.


\bibliography{main}
\bibliographystyle{plain}

\end{document}